\documentclass[letterpaper, 10 pt, conference]{ieeeconf}  

\IEEEoverridecommandlockouts                              

\usepackage{graphics} 
\usepackage{epsfig} 
\usepackage{amsmath} 
\usepackage{amssymb}  

\usepackage{multirow}
\usepackage{verbatim}
\usepackage{amsfonts}    
\usepackage{booktabs}
\usepackage{algorithm}   
\usepackage[noend]{algpseudocode} 

\floatname{algorithm}{Procedure}

\usepackage{xcolor}
\usepackage{stfloats}

\makeatletter
\let\NAT@parse\undefined
\makeatother
\usepackage{hyperref}
\hypersetup{colorlinks,citecolor={blue}}

\title{\Large SAFER: Safe Collision Avoidance using Focused and Efficient Trajectory Search with Reinforcement Learning}

\author{Mario Srouji, Hugues Thomas, Yao-Hung Hubert Tsai, Ali Farhadi, Jian Zhang \\ Apple Inc., 
  \{msrouji,hthomas23,yaohung\_tsai,afarhadi,jianz\}@apple.com
\thanks{Accepted in IEEE International Conference on Automation Science and Engineering (CASE), 2023.} 
}

\begin{document}

\maketitle
\thispagestyle{empty}
\pagestyle{empty}

\begin{abstract}
Collision avoidance is key for mobile robots and agents to operate safely in the real world. In this work we present SAFER, an efficient and effective collision avoidance system 
that is able to improve safety by correcting the control commands sent by an operator.
It combines real-world reinforcement learning (RL), search-based online trajectory planning, and automatic emergency intervention, e.g. automatic emergency braking (AEB). The goal of the RL is to learn an effective corrective control action that is used in a focused search for collision-free trajectories, and to reduce the frequency of triggering automatic emergency braking. This novel setup enables the RL policy to learn safely and directly on mobile robots in a real-world indoor environment, minimizing actual crashes even during training. Our real-world experiments show that, when compared with several baselines, our approach enjoys a higher average speed, lower crash rate, less emergency intervention, smaller computation overhead, and smoother overall control.

\end{abstract}

\section{Introduction} 

Mobile robots are slowly but surely taking a place in our everyday lives and work environments with various applications: vacuum cleaning, video recording, companionship, security, tele-presence, etc. Whether they are autonomous agents or controlled by human operators, collision avoidance is key for mobile agents to operate safely, and effectively in the real world. There are numerous approaches to collision avoidance, including search-based planning methods, trajectory optimization, learning-based methods, and emergency intervention systems. Search-based trajectory planning methods are successful at finding collision-free trajectories if given good discretization, enough computation, and ideal search heuristics, however, due to the size of the search space in real-world continuous problems, the search could be too computationally heavy to yield good enough solutions ~\cite{fox1997dynamic,likhachev2002lifelong,vsvec2014target,shah2016resolution,8794192}. Trajectory optimization is able to solve for locally optimal trajectories for collision avoidance, however, it requires very good initialization, and sophisticated constraint modeling to guarantee continuity, and feasibility ~\cite{xiaojing2020optimization,rosmann2017integrated,7525254}. Learning-based methods are promising as they are data-driven, and run inference on a GPU or AI accelerator to achieve fast and fixed computation, however, it is difficult to ensure safety due to distribution shift and uncertainties, especially during training~\cite{gandhi2017learning,kahn2021badgr}. Another approach is automatic emergency intervention system, e.g. automatic emergency braking (AEB), which sacrifices optimality in exchange for fast computation and low false negative rate, by bringing the agent to a complete stop in an emergency~\cite{kahn2017uncertainty,hulshof2013autonomous,rosen2013autonomous}.

The AEB is activated when the agent has encountered an unsafe state that requires immediate takeover. Our key insight is that even though this unsafe state could be a false positive in certain situations, it can be treated as a conservative signal for imminent collision. Our approach treats this conservative signal as a learning signal, to encourage reducing its frequency of activation, and avoid stop-and-go behaviors. Specifically, we train an RL policy to minimize the number of AEB activation, and improve collision avoidance metrics. The policy outputs corrective control commands that are then refined with the Dynamic Window Approach (DWA) method~\cite{fox1997dynamic} using a reduced search space. These corrective control commands correct an agent's upstream control (e.g from human control or other upstream algorithms) when detecting potential collisions.

\begin{figure}[b!]
    \vspace{-4mm}
    \centering
    \includegraphics[width=0.98\columnwidth]{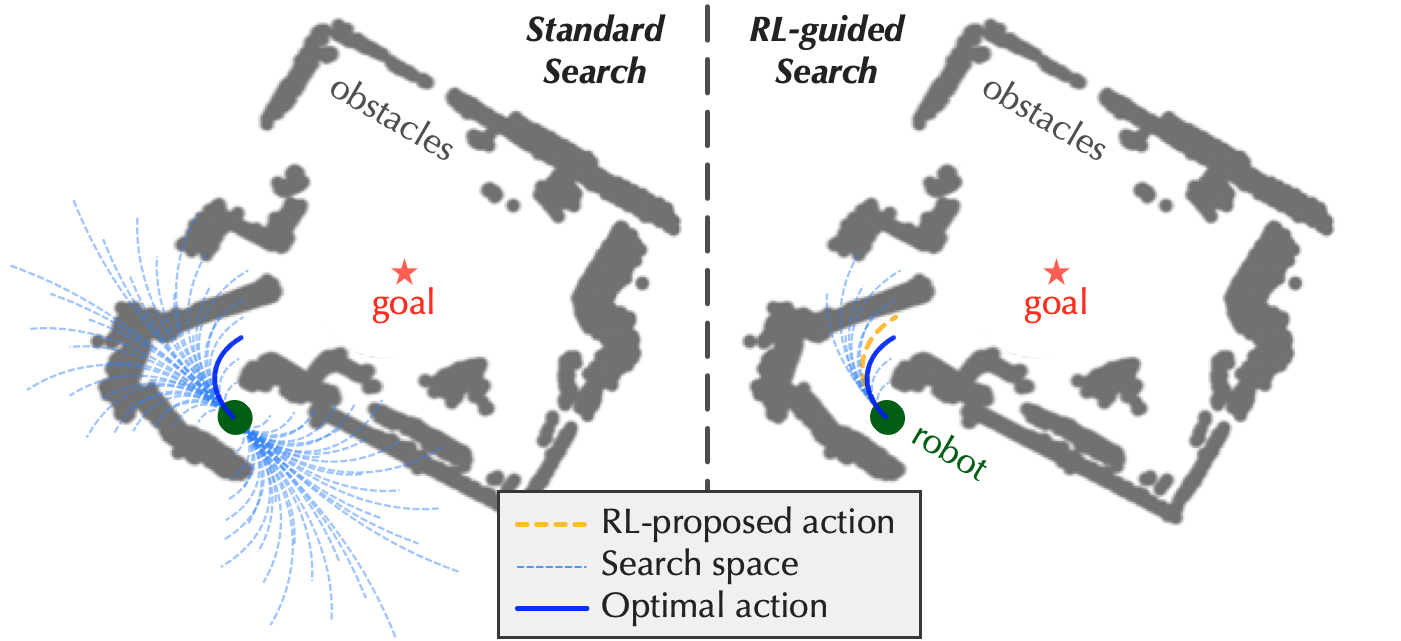}
    \vspace{-2mm}
    \caption{Our method uses a Reinforcement Learning (RL) policy to propose an initial action. Using this initial action, we reduce the search space to find the optimal action faster.}
    \label{fig:rl-guided}
    \vspace{0mm}
\end{figure}

Our collision avoidance system SAFER takes input from lidar and ultrasonic sensor scans, wheel odometry for robot state, and the upstream control commands. We fuse the lidar and ultrasonic sensor scans to detect a diverse set of obstacles, including transparent glass, reflective surfaces, furniture, humans, etc. 
We design a reward function for our RL agent with two terms. The first term encourages the reduction of AEB activation. The second term improves collision avoidance metrics through a cost function, such as average speed, distance to obstacles, and matching human control intention. Our focused DWA search introduces additional parameters to define the reduced search space around the output of the RL policy. This focused search space, shown in Fig. \ref{fig:rl-guided}, reduces the latency of our method, which is crucial for a reactive policy in a dynamic environment.

Our SAFER method learns directly in a real-world indoor office environment by using a distributed RL training setup, leveraging the Soft Actor Critic (SAC)~\cite{haarnoja2018soft} algorithm. Our robot collects experience through a human driver attempting to follow pre-defined routes, as well as intentionally trying to crash the robot. A centralized training server collects experiences and updates the RL policy. With our novel setup, we can perform real-world RL training while minimizing actual crashes throughout the learning process, avoiding physical damage. We believe this work opens up a new approach for safer learning in the real world, and thus accelerates the development of intelligent agents.

\begin{figure*}[t!]
    \vspace{0mm}
    \centering
    \includegraphics[width=0.9\textwidth]{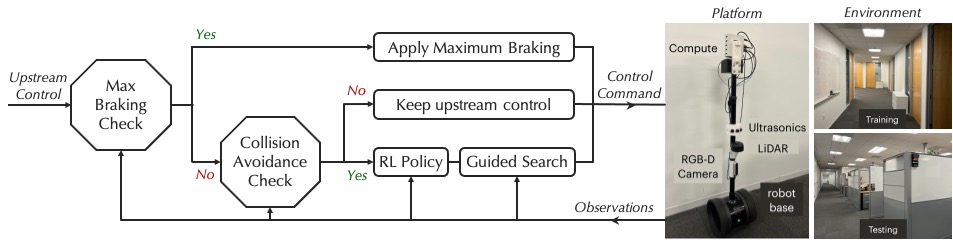}
    \vspace{-2mm}
    \caption{Diagram of our SAFER collision avoidance system, next to illustrations of our robotic platform and our training and test environments. Our differential drive self-balancing robot is based off of a headless Double3, and has an onboard Linux computer (Nvidia Jetson NX), a 2D $360^\circ$ lidar scanner (RPLidar A3), and an ultrasonic sensor array (Maxbotix MB-1040). The RGB-D cameras pictured are not used in this work.}
    \label{fig:mainfigure}
    \vspace{-5mm}
\end{figure*}

\vspace{-1mm}
\section{Related Work}
\vspace{-1mm}

There's a broad literature on the topic of real-world collision-avoidance, and in the following we discuss its related work on several topics.

{\bf Safety in Real-world Robots.} AEB is commonly applied in real-world robots, where AEB performs maximum braking when detecting obstacles or emergency situations. Nonetheless, the AEB system could make sub-optimal decisions such as performing maximum braking too late, which can result in a head-on collision. Therefore, it's safer to plan ahead to avoid getting into a situation where an AEB system has to take over~\cite{fox1997dynamic}, by formulating collision avoidance as one of the cost terms in an overall search or trajectory optimization problem~\cite{kahn2018self,xiaojing2020optimization}. 
With enough computational power, model simplification and heuristics, such approaches have hope for finding collision-free global optimal solutions. Yet, the computational resources on a mobile robot are often limited, and there are other essential tasks that require computation like perception \cite{lavalle2006planning}. Our work considers a realistic setup where we perform efficient collision avoidance using on-device compute to prevent the robot from getting into an unsafe position. It relates to safe reinforcement learning, such as constrained policy optimization (CPO)~\cite{achiam2017constrained}. While there is no evidence that shows that CPO works in a real world system, our approach trains and operates in the real world. Other works leverage RL with the goal of designing controllers that are probabilistically-safe using verification mechanisms~\cite{bozkurt2020control,hasanbeig2019reinforcement}, and shielding methods~\cite{alshiekh2018safe,elsayed2021safe,bastani2021safe}. Lastly, there are some works that assume the dynamics of the real-world are known~\cite{van2008reciprocal,otte2016rrtx,zhu2021learning}. Our approach, on the other hand, does not have this assumption.

{\bf Learning from Human Demonstrations.} A popular way to learn to avoid collisions is to learn from human demonstrations, including imitation learning methods~\cite{muller2005off,pan2017agile,codevilla2018end,ross2013learning,giusti2015machine,loquercio2018dronet}, and by learning to avoid human disengagement~\cite{kahn2021land}. The first class of methods (imitation learning) aim to make a robotic agent mimic human behaviors. The second class of methods (learning to avoid human disengagement) aim to make a robotic agent generate actions that avoid human disengagement, which is similar to our approach where we learn to reduce the amount of AEB activation. Although these approaches have demonstrated their effectiveness in collision avoidance, they require a lot of human demonstration data, which is expensive to collect. Our approach on the other hand, learns from AEB activation signals, which require no human demonstrations.

{\bf Sim-to-Real Transfer.} Another family of collision avoidance involves initially learning in simulation, and then performing sim-to-real transfer~\cite{sadeghi2016cad2rl,muller2018driving,hirose2019deep,wenzel2021vision,zhao2020sim}. Although simulation offers a large amount of training data when compared to the real-world, the sim-to-real domain gap may hinder the learning algorithms to generalize. Our approach learns directly in the real world, and evidence shows that direct real world learning is preferable to applying sim-to-real transfer~\cite{kirk2021survey}.

{\bf Learning from Collisions.} Our work also relates to learning collision avoidance by experiencing collisions~\cite{kahn2017uncertainty,gandhi2017learning,kahn2021badgr}, in which collision is regarded to be a valuable learning signal. However these approaches will inevitably cause physical damage to robotic agents, and the environment in which they operate in. On the other hand, our approach regards AEB activation as a valuable learning signal, which is safer in the real world.   

{\bf Combining RL and DWA.} There are previous works that relate to combining RL with the dynamic windows approach (DWA). Some prior works learn to adaptively change the weight coefficients of the DWA evaluation function using RL, or extend them to provide better path planning and obstacle avoidance \cite{heo2021dynamic, dobrevski2020adaptive, chang2021reinforcement, patel2020dynamically}, while others leverage RL to learn a velocity model for objects to avoid obstacles \cite{kim2022improvement}. The difference between our work, and these prior works is that we leverage RL to improve the efficiency and granularity of the search in DWA. 
\section{Problem Formulation}
We make certain assumptions in our work in order to simplify the real world collision avoidance. Our environment is assumed to be composed of static obstacles, i.e we do not perform tracking for humans or other dynamic obstacles. We also assume the robot's actuation limitations are measured and known ahead of time, and we fine-tune a kinematic model for our environment to make trajectory predictions, which also assume a constant velocity model. In the following, we introduce some important notation and parameters that should be referenced in the rest of the paper:

\underline{Robot Kinematics}
\begin{itemize}
     \item {Reaction time $t_r=100ms$} represents our collision avoidance module's latency, and is a measured quantity.
     \item {Robot states $\mathbf{x}=(x, y, \theta, v, \omega)$} with $x$/$y$ being 2D euclidean coordinates, $\theta$ being the robot's yaw, and $v$/$\omega$ being the robot's linear velocity/angular velocity.
     \item {Maximum robot linear \& angular acceleration $(a_{\rm max}^{v}$, $a_{\rm max}^{\omega})$.}
     \item {Maximum robot linear \& angular velocity ($v_{\rm max}$, $\omega_{\rm max}$). Minimum velocities are assumed to be the negative of the maximum}.
     \item {Robot kinematics $\mathbf{x}_{i+1} := \operatorname{f}(\mathbf{x}_i)$} is defined as
    \begin{equation*}
    \small
    {\mathbf{x}_{i+1} :=  \operatorname{f}(\mathbf{x}_i)}
    = \left\{\begin{matrix}
    x_{i+1} \,\,\, = & x_i + v_i\cos(\theta_i) t_r \\
    y_{i+1}  \,\,\, = & y_i + v_i\sin(\theta_i) t_r\\
    \theta_{i+1} \,\,\, = & \theta_i + \omega_i t_r \\
    v_{i+1} \,\,\, = & v_c \\
    \omega_{i+1} \,\,\, = & \omega_c
    \end{matrix}\right. .
    \label{eq:robot_dynamics}
    \end{equation*}
    \item  {Robot trajectory $\mathbf{traj}(v, \omega, \Delta t)$} is defined as
    \begin{equation*}
    \small
    \mathbf{traj}(v, \omega, \Delta t) :=[\mathbf{x}_0, \mathbf{x}_1, \mathbf{x}_2, ..., \mathbf{x}_{\Delta t / t_r}],
    \label{eq:robot_trajectory}
    \end{equation*}
    where $t_r$ is considered to be the time step and $\Delta t$ is the total time horizon. The initial robot state $\mathbf{x}_0=(0, 0, 0, v, \omega)$ since we use an ego-centric robot frame. 
\end{itemize}

\underline{Sensor and Obstacles}
\begin{itemize}
    \item {2D Lidar scan $\{l_0, l_1, \cdots, l_{359}\}$} collects signals from $360^\circ$, representing distance in meters.
    \item {2D ultrasonic scans $\{u_{-45}, u_0, u_{45}\}$} collects signals from $\{-45^\circ, 0^\circ, 45^\circ\}$, representing distance in meters. We leverage ultrasonics to detect glass surfaces. 
    \item {Obstacles $\mathbb{O}_s\subset\mathbb{R}^2$} are registered using lidar and ultrasonic signals $\{l_0, l_1, \cdots, l_{356}, u_{-45}, u_0, u_{45}\}$ within $t_r$.
\end{itemize}

\underline{Collision Avoidance Parameters}
\begin{itemize}
    \item {$\sigma$ is an indicator variable} and is $=1$ when maximum braking is triggered in the following time step; otherwise $=0$.
    \item {Reference control commands $(v_{ref}, \omega_{ref})$} are from human control input or other upstream control.
    \item {Corrective control commands $(v_{c}, \omega_{c})$} are outputted by our collision avoidance module.
    \item {Plan-ahead time $t_p$} is a hyper-parameter, representing the time horizon for trajectory planning, according to the maximum braking capability of the robot. $t_p = t_r + v/(2 a_{\rm max}^v)$. 
 \end{itemize}

\underline{Tune-able Hyperparameters}
\begin{itemize}
    \item $\beta > 1$ is a constant expanding the plan-ahead time for collision avoidance prediction to $\beta t_p$.
    \item $n_v, v_\omega$ control the amount of linear and angular velocity samples, respectively, used in standard DWA trajectory search.
    \item $\lambda_1, \lambda_2$ are constants in the RL reward function.
    \item $\gamma \in (0, 1)$ controls the size of the RL-guided search window, in comparison to the standard window in DWA.
    \item $\delta \in (0, 1)$ controls the number of trajectories that are sampled in the RL-guided search window, in comparison to the standard number of DWA samples $n_v, v_\omega$.
\end{itemize}
\vspace{-1mm}
\section{SAFER Method}
\vspace{-1mm}

We present our collision avoidance system in Fig.~\ref{fig:mainfigure} (a), noting that our method runs in parallel with upstream tasks, such as navigation planning or human control. Our approach takes as input the robot state from odometry, the lidar and ultrasonic scans, and the control command from the upstream tasks. The output is a corrective control command to the upstream reference tasks, which avoids collision and provides favorable behavior (fast speed, smooth control, etc.). There are three stages within our collision avoidance: {\em maximum braking and collision avoidance checks} (details in Sec.~\ref{subsec:safety_check}), {\em RL collision avoidance policy} (details in Sec.~\ref{subsec:RL_policy_search}), {and {\em focused DWA search} (details in Sec.~\ref{subsec:focused_DWA_search})}. In the first stage, we determine whether the robot requires maximum braking, or a corrective control command through our RL policy. In the second stage, we apply our RL policy to output corrective control commands with the goal of 1) reducing the frequency of emergency braking interventions and 2) avoiding collision effectively. In the third stage, we use a DWA search focused on a small space around the RL policy output to find the best corrective control command. In the final section Sec.~\ref{subsec:hyperparameter} we explain how to tune important hyperparameters in our SAFER method.


\subsection{Maximum Braking and Collision Avoidance}
\label{subsec:safety_check}
  According to the robot's current state and the surrounding obstacles, we first check whether emergency intervention or collision avoidance is necessary, and apply corrective action to the upstream control commands. If emergency intervention is required, we apply maximum braking to bring the robot to a complete stop. If emergency braking is not required, but collision avoidance is, we apply a corrective control command using our RL policy. If neither emergency braking nor collision avoidance is required, our system maintains the upstream control.
 
 \underline{Collision.} We define $\mathbb{E}(\mathbf{x})\in\mathbb{R}^2$ as the set of points in 2D space by the robot's shape at state $\mathbf{x}$. A collision occurs between the robot's trajectory $\mathbf{traj}(v, \omega, \Delta t)$ and the set of obstacles $\mathbb{O}_s\subset\mathbb{R}^2$ if:
\begin{equation}
    \bigcup_{\mathbf{x}\in \mathbf{traj}(v, \omega, \Delta t)}(\mathbb{E}(\mathbf{x}) \cap \mathbb{O}_s) \neq \emptyset.
\label{eq:collision}
\end{equation}
 
 \underline{Emergency Braking} We adopt maximum braking control when the robot's trajectory along the plan-ahead time $t_p$, denoted as $\mathbf{traj}(v, \omega, t_p)$, collides with obstacles, formulated as {\small $\bigcup_{\mathbf{x}\in \mathbf{traj}(v, \omega, t_p)}\mathbb{E}(\mathbf{x}) \cap \mathbb{O}_s \neq \emptyset$}.
 
 \underline{Collision Avoidance} If maximum braking is not required, then we expand the plan-ahead time to $\beta t_p > 1$\footnote{We found $\beta = 2$ to be a good balance to avoid emergency braking, and minimize collision avoidance intervention to upstream control.} to see if the robot collides with obstacles that are further away. It can be formulated as {\small $\bigcup_{\mathbf{x}\in \mathbf{traj}(v, \omega, \beta t_p)}\mathbb{E}(\mathbf{x}) \cap \mathbb{O}_s \neq \emptyset$}. If there is a collision, then we use our RL policy to output a corrective control command.


\subsection{RL Collision Avoidance Policy}
\label{subsec:RL_policy_search}
\begin{figure}[t!]
    \vspace{0mm}
    \centering
    \includegraphics[width=0.8\columnwidth]{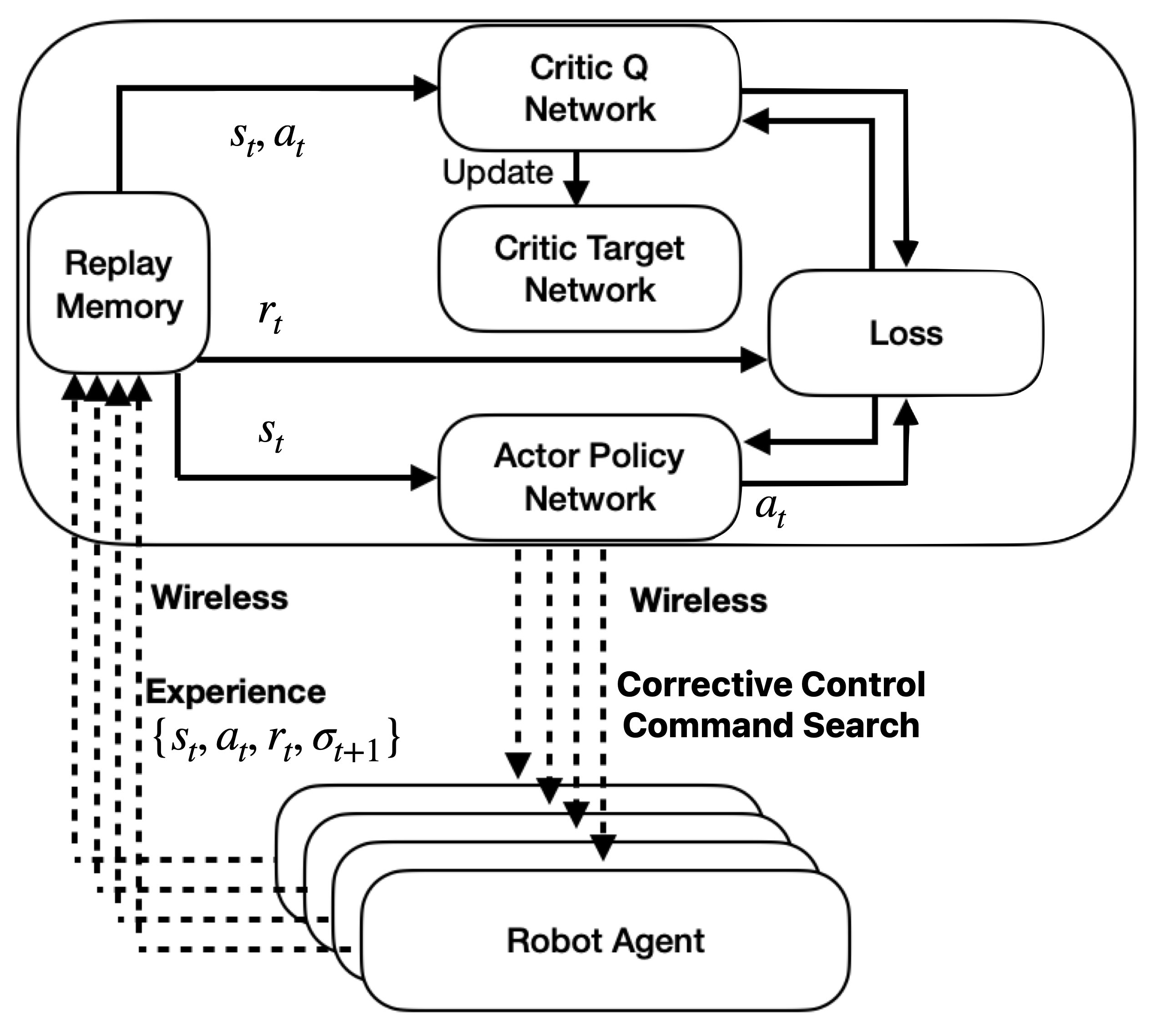}
    \includegraphics[width=0.75\linewidth]{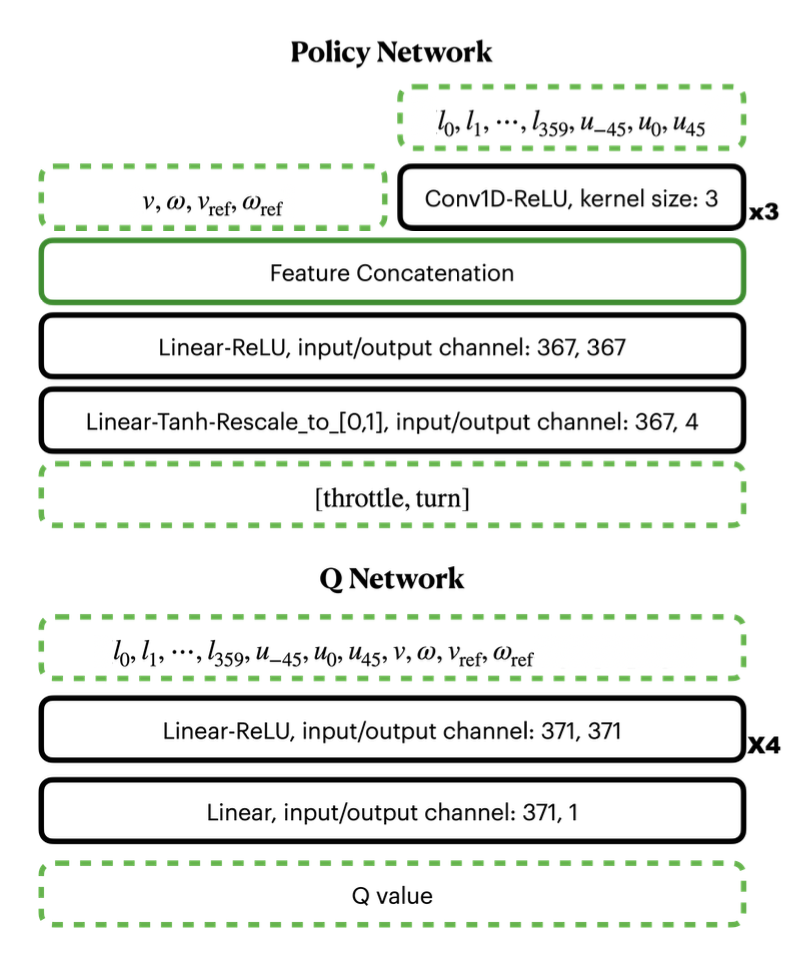}
    \vspace{0mm}
    \caption{Our distributed RL training setup (top) and RL network.}
    \label{fig:rl_policy}
    \vspace{-4mm}
\end{figure}

We consider a distributed RL training setup with the soft actor critic (SAC)~\cite{haarnoja2018soft} approach. Our robots collect experiences and send them to a central training server, and the training server uses these experiences to train and update the actor's policy network, as well as the critic and critic target networks. The training server sends all robots the updated actor network after a fixed number of training steps. Fig.~\ref{fig:rl_policy} presents the overall setup and network specifications.


For each time step, $s_t$ represents the input to the policy and the critic network, consisting of lidar and ultrasonic scans, robot's linear and angular velocity, and the control commands from the upstream task: $s_t = [l_0, l_1, \cdots, l_{359}, u_{-45}, u_0, u_{45}, v, \omega, v_{\rm ref}, \omega_{\rm ref}]$. 

$a_t$ is the output of the Policy Network and also part of the input to the critic's Q Network: $a_t = [throttle, turn] \in [-1, 1]$

$r_t$ is our reward function. Since we aim to learn corrective control commands that can 1) reduce the number of maximum braking interventions and 2) avoid collision effectively, we compose $r_t$ to be
\begin{equation}
\small
r_t = -\lambda_1 \sigma_{t+1} - \lambda_2 J(v_c, \omega_c),  
\label{eq:reward}
\end{equation}
where $\lambda_1, \lambda_2$ are hyper-parameters (in our design, $\lambda_1=35, \lambda_2=10$), $\sigma_{t+1}$ indicates whether maximum braking was performed in the next time step, and $J(v_c, \omega_c)$ is the cost of the control command $(v_c, \omega_c)$. Precisely, we define $J(v_c, \omega_c)$ as
\begin{equation}
\small
\begin{aligned}
   J(v_c, \omega_c) &= \mathrm{c}_1(v_{\rm max} - v_c) + \mathrm{c}_2 (|v_c-v_{\rm ref}| + |\omega_c-\omega_{\rm ref}|)\\
   & \quad+ \mathrm{c}_3\frac{1}{\operatorname{dist}\big(\mathbb{O}_s, \mathbf{traj}(v_c, \omega_c, \beta t_p)\big)},
\label{eq:cost}
\end{aligned}
\end{equation}
where $c_1, c_2, c_3$ are hyper-parameters (in our settings, $c_1=0.4, c_2=0.4, c_3=0.2$).  $(v_{\rm max} - v_c)$ encourages the corrective linear velocity to be fast, $(|v_c-v_{\rm ref}| + |\omega_c-\omega_{\rm ref}|)$ minimizes the deviation of corrective control commands from upstream control commands, and $\frac{1}{\rm dist(\cdot)}$ encourages the robot to stay away from obstacles. We define $\operatorname{dist}(\mathbb{O}_s, \mathbf{traj})$ to be the distance between the robot's trajectory and the obstacles:
{\small $$
 \operatorname{dist}(\mathbb{O}_s, \mathbf{traj}) = \min_{\forall (x_s, y_s) \in \mathbb{O}_s, (\hat{x}, \hat{y})\in\mathbf{x}\in\mathbf{traj}} \lVert (\hat{x}, \hat{y}) - (x_s, y_s) \rVert .
$$}

\subsection{Focused DWA Search}
\label{subsec:focused_DWA_search}
The dynamic window approach (DWA)~\cite{fox1997dynamic} defines a dynamic window $W=[v_{\rm lower}, v_{\rm upper}] \times [\omega_{\rm lower}, \omega_{\rm upper}]$ identifying the set formed by the Cartesian product of values in the range between lower and upper linear velocity, and angular velocity values. DWA then performs an exhaustive search for a control command $(v_c, \omega_c)$ $\in$ $W$ that minimizes a defined cost. Alternatively, our approach improves upon this by using the output of an RL policy to guide DWA, allowing us to perform a smaller targeted search to find the optimal control. 

\underline{Standard Dynamic Window.} The standard dynamic window $W$ is defined using the current robot velocity $(v,\omega)$, the maximum linear and angular acceleration $(a_{\rm max}^v, a_{\rm max}^\omega)$, and the robot reaction-time $t_r$:
\begin{equation}
\begin{aligned}
W = [max(v-a_{\rm max}^vt_r, -v_{\rm max}), min(v+a_{\rm max}^vt_r, v_{\rm max})] \times 
\\
[max(\omega-a_{\rm max}^\omega t_r, -\omega_{\rm max}), min(\omega+a_{\rm max}^\omega t_r, \omega_{\rm max})]. 
\label{eq:window}
\end{aligned}
\end{equation}
Searching in $W$ can be inefficient since the ranges between the lower and upper velocities can be large, and small search granularity is needed to find good solutions. In DWA we typically consider a search consisting of $n_v$ linear and $n_\omega$ angular velocity control commands inside of the window $W$, which yields a search space of $n_v*n_\omega$ trajectories. The granularity of such a search (at maximum) is $2a_{\rm max}^vt_r / n_v$ for linear velocity, and $2a_{\rm max}^\omega t_r / n_\omega$ for angular velocity.

\underline{RL-guided Search.} After we obtain $throttle, turn$ from the RL policy network, we scale the output as $[v_s, \omega_s]=[throttle*v_{\rm max}, turn*\omega_{\rm max}]$. We then check to make sure that the scaled output $[v_s, \omega_s]$ is feasible and lies within the dynamic window $W=[v_{\rm lower}, v_{\rm upper}] \times [\omega_{\rm lower}, \omega_{\rm upper}]$ as defined in Eq.~\ref{eq:window}. If $[v_s, \omega_s] \notin W$, then we threshold the scaled output to fit into $W$:
\begin{equation}
\small
[max(min(v_s, v_{\rm upper}), v_{\rm lower}), max(min(\omega_s, \omega_{\rm upper}), \omega_{\rm lower})]
\label{eq:rlcommand}
\end{equation}
Next we perform a focused search around the RL action using parameters $\gamma \in (0, 1)$ and $\delta \in (0, 1)$ inspired by the traditional DWA window discussed in Eq.~\ref{eq:window} ($\gamma=0.05$ and $\delta=0.1$ in our experiments). Our search picks the corrective control $v_c, \omega_c$ that minimizes $J(v_c, \omega_c)$ in Eq.~\ref{eq:cost}. We create a search window around $[v_s, \omega_s]$ that is $\gamma$ times the size of the standard window in Eq.~\ref{eq:window}. Our $[v_{\rm lower}, v_{\rm upper}] \times [\omega_{\rm lower}, \omega_{\rm upper}]$ becomes:
\begin{equation}
\begin{aligned}
[max(v_s-\gamma a_{\rm max}^vt_r, -v_{\rm max}), min(v_s+\gamma a_{\rm max}^vt_r, v_{\rm max})] \times 
\\
[max(\omega_s-\gamma a_{\rm max}^\omega t_r, -\omega_{\rm max}), min(\omega_s+ \gamma a_{\rm max}^\omega t_r, \omega_{\rm max})]. 
\label{eq:reduced_window}
\end{aligned}
\end{equation}
We then sample $\delta n_v$ linear velocities and $\delta n_\omega$ angular velocities from the smaller window, resulting in a total of $\delta^2 n_v n_\omega$ trajectories being searched. This also yields a search granularity (at maximum) of $2\gamma a_{\rm max}^vt_r / \delta n_v$ for linear velocity, and $2\gamma a_{\rm max}^\omega t_r / \delta n_\omega$ for angular velocity, meaning the granularity is $\gamma / \delta$ compared to standard. To summarize, our targeted search window is $1 / \gamma$ times smaller, and we sample $1 / \delta^2$ less trajectories.

\subsection{Tuning Hyperparameters}
\label{subsec:hyperparameter}
In order to aid the reader in understanding how to tune the more important collision avoidance parameters in our work, $[\beta, n_v, n_\omega, \gamma, \delta]$, we provide a brief discussion. 
\begin{itemize}
    \item $\beta>1$ controls the planning look-ahead multiplier. Larger settings cause the RL-guided search to be triggered more often for collision avoidance. Too large of a setting can negatively affect the human control experience due to being overly conservative. We set $\beta=2$.
    \item $n_v, n_\omega$ determine how many linear and angular velocities are sampled from the dynamic window, resulting in $n_v n_\omega$ samples. We maximize the values of $n_v, n_\omega$ up to the point where the standard DWA search has a latency of $t_r$, which is 100ms in our setting, and can be tuned as required for the robotic platform. One can increase the value of $n_v$ compared to $n_\omega$ or vice-versa if the window is larger in either dimension. 
    \item $\gamma, \delta \in (0, 1)$ control the size of the targeted RL search. $\delta$ should be set such that the search time $t_{search}$ resulting from the number of samples $\delta^2 n_v n_\omega$ offsets the inference time of the RL policy $t_{\pi}$; $t_{search} \leq t_r - t_{\pi}$. $\gamma$ is then set to achieve the desired search granularity compared to the standard DWA search $\gamma / \delta$, ideally $\gamma / \delta \leq 1$ (more granular). 
\end{itemize}


\vspace{-1mm}
\section{Experiments}

\vspace{-1mm}
In this section, we perform real-world experiments by controlling a robot in an indoor office environment containing a variety of obstacles including glass, humans, office furniture, doors, hallways, etc. We perform our tests in a new part of the environment unseen during the training phase, with a different office layout and obstacles as shown in Fig.~\ref{fig:mainfigure}. First, we leverage a sinusoidal driving policy which removes any human bias in the results. Then we let a human drive the robot with specific instructions. The human operator tries to follow pre-determined (marked) routes through corridors, doorways, around furniture, etc. We also perform experiments where the human operator intentionally tries to collide the robot with obstacles in the environment for a fixed amount of time. Finally, we show qualitative comparisons between our SAFER method's collision avoidance compared to standard DWA search.

\begin{table}[t!]
\vspace{0mm}
\centering
\caption{\small Details of the compared methods.}
\vspace{-2mm}
\scalebox{1.0}{ \begin{tabular}{c|cccc}
\toprule
\multirow{2}{*}{Methods} & \multirow{2}{*}{Search} &  Learning & \multirow{2}{*}{latency ($\downarrow$)} & search \\ & & from AEB & & size ($\downarrow$) \\
\midrule
NoSafety  & no &  no &  n/a & n/a
\\ 
AEB & no & no & 8ms & n/a
\\ 
DWA & yes & no & 100ms & 2500
\\ 
RL  & no & yes & 15ms  & n/a
\\
SAFER (ours) & yes & yes & 20ms & 25
\\ \bottomrule
\end{tabular}
}
\label{tbl:baselines}
\vspace{-5mm}
\end{table}

\underline{Methods of Comparison.} We consider five baselines to demonstrate the effectiveness of our method, all of which are developed by us as separate parts of the SAFER method. The first baseline only considers human control, without a collision avoidance system. Building upon the first baseline, the second, third, fourth, and fifth baselines add a collision avoidance component to correct the human control (same human driver). For the second baseline, the collision avoidance only considers automatic emergency maximum braking. The subsequent baselines add a plan-ahead collision avoidance mechanism in addition to the automatic emergency braking. The third baseline is the most similar to our approach, by leveraging emergency maximum braking, and a plan-ahead collision avoidance policy using standard DWA search as described in the method section. The fourth baseline replaces the standard DWA search in the plan-ahead collision avoidance system with our trained RL policy, but without the focused RL-guided DWA search. The fifth baseline is our SAFER approach, which adds the RL-guided focused search to the output of the RL policy for collision avoidance. We shorthand the first baseline as $\mathsf{NoSafety}$, the second baseline as $\mathsf{AEB}$, the third baseline as $\mathsf{DWA}$, the fourth baseline as $\mathsf{RL}$, and our method as $\mathsf{SAFER}$. Note that the RL policy in $\mathsf{RL}$ and $\mathsf{SAFER}$ is identical, and the methods only differ in that $\mathsf{SAFER}$ leverages the focused search around the output of the RL.

\underline{Metrics.} We report the following metrics which should be maximized ($\uparrow$) or minimized ($\downarrow$): 

\begin{itemize}
  \item $\mathsf{successes}$
  ($\uparrow$): the number of successful trials as detailed in Sec. \ref{subsec:sinus}.
  \item $\mathsf{collisions}$
  ($\downarrow$): the number of actual collisions during a driving session or across multiple trials.
  \item $\mathsf{average\,speed}$ ($\uparrow$): the mean of the linear speed across all control decisions in ($m/s$).
  \item $\mathsf{\#\,max\,braking}$
  ($\downarrow$): the number of times maximum braking was performed during a driving session.
  \item $\mathsf{latency}$
  ($\downarrow$): the total resulting collision avoidance planning cycle time.
  \item $\mathsf{unsmoothness}$
  ($\downarrow$): the mean of the linear and lateral accelerations between control decisions. As an example, given two sequential control decisions $(v_0, w_0)$ and $(v_1, w_1)$, and elapsed time of $t$ we calculate $\mathsf{unsmoothness}$ for this step to be $((v_1-v_0)^2 + (w_1-w_0)^2)/t$.
  \item $\mathsf{action cost}$
  ($\downarrow$): the cost of the control action according to Eq.~\ref{eq:cost}.
\end{itemize}

\begin{table}[t!]
\vspace{0mm}
\centering
\caption{\small 30 Trials: Sinusoidal Policy in Tight Doorway (static) and Human Encounter (dynamic) scenarios.}
\vspace{0mm}
\scalebox{1.0}{ \begin{tabular}{c|cc|cc}
\toprule

\multirow{2}{*}{Methods} & \multicolumn{2}{c|}{Tight Doorway} & \multicolumn{2}{c}{Human Encounter} 
\\ 
& succ. ($\uparrow$) & colli. ($\downarrow$) & succ. ($\uparrow$) & colli. ($\downarrow$) \\
\midrule
AEB & 0 & 3 & 0 & 9
\\ 
DWA & 23 & 2 & 22 & 3      
\\ 
RL & 19 & 5 & 23 & 6
\\
SAFER (ours) & 27 & 0 & 28 & 1           
\\ \bottomrule
\end{tabular}
}
\label{tbl:fixedexp}
\vspace{-5mm}
\end{table}

\begin{table*}[b!]
\vspace{-4mm}
\centering
\caption{\small Expert Human Operator Experiments.}
\vspace{-2mm}
\scalebox{1.0}{ \begin{tabular}{cc|ccccc}
\toprule
 Operator & \multirow{2}{*}{Methods} & average  & collisions ($\downarrow$) & \# max & unsmoothness ($\downarrow$) & avg. action \\ instructions & & speed ($\uparrow$) & & braking ($\downarrow$) & & cost ($\downarrow$)\\
\midrule
 & NoSafety  &       0.47 m/s        &   14   &  n/a    &    0.23  & 0.41
\\ 
Follow a  & AEB &       0.31 m/s        &    4    & 27   &     0.31   & 0.32 
\\ 
predefined route (1hr.) & DWA &        0.44 m/s        &   2   & 7      &     0.19     & 0.24 
\\ 
 & RL  &      \textbf{0.53 m/s}        &    3  &  8       &      0.27     & 0.29 
\\
 & SAFER (ours) &  0.51 m/s       &  \textbf{1}   & \textbf{3}    &   \textbf{0.12}   & \textbf{0.16}                       
\\
\midrule
 & AEB & 0.31 m/s & 5 & 19 & 0.37 &  0.47     
\\ 
Try to  & DWA & 0.43 m/s & 2 & 7 & 0.22 &  0.33         
\\ 
crash (10mins.) & RL  & 0.39 m/s & 3 & 9 & 0.49 &  0.35    
\\ 
 & SAFER (ours) & \textbf{0.50 m/s} & \textbf{0} & \textbf{3} & \textbf{0.14} & \textbf{0.21}       
\\ \bottomrule
\end{tabular}
}
\label{tbl:humanexp}
\vspace{0mm}
\end{table*}

\subsection{Sinusoidal Driving Policy in Hard Cases}
\label{subsec:sinus}
Our first experiment aims at evaluating our method in a controlled setup, without any bias from a human operator. We define two challenging scenarios involving static and dynamic obstacles. In the first scenario (Tight Doorway), the robot approaches a partially closed doorway, requiring a tight turn to enter the room, or a larger turn to avoid the entrance altogether. We consider a success to be the robot entering the room without colliding with any object.The second scenario (Human Encounter), involves the robot driving down a hallway, where a human walks in the opposite direction in a straight line directly toward the robot. We start the human at the same marked location each time. Success in this case requires the robot to turn away from the human, and continue making progress down the hallway in the same original direction without collision. 

For both scenarios, the robot is controlled by our sinusoidal policy, which sets forward throttle to a constant maximum, while turn varies according to time in seconds $t$; $turn = sin(t)$. We mark a baseline starting location for each scenario (position and angle) and set the actual starting angle $\theta$ for each run according to a random uniform distribution $\theta + \frac{\pi}{4} * \mathcal{U}(-1, 1)$. In Table~\ref{tbl:fixedexp}, we report the number of real collisions and successes across 30 trials for each baseline and both scenarios. Our $\mathsf{SAFER}$ method has the highest number of successes, and lowest number of collisions in both scenarios when compared to the baselines.

\subsection{Human Operator Results} 
\vspace{-1mm}
We conducted experiments to validate that our SAFER method behaves as intended in real-life scenarios. Our next two setups thus involve the robot being controlled by an expert human operator (hours of robot driving experience), where the operator is asked to follow specific instructions in each setup. Firstly, we ask the operator to drive the robot continuously for one hour through pre-determined routes in our indoor office environment. We chose difficult routes that require navigating around furniture, traversing doorways/hallways, and other challenging obstacles such as glass, boxes, etc. while in the presence of humans. Secondly, the operator is tasked with intentionally crashing the robot randomly in the evaluation environment. The results are compiled in Table~\ref{tbl:humanexp}.

\underline{Following Pre-Determined Routes}
Our experiments reveal that implementing a simple mechanism like $\mathsf{AEB}$ reduces actual collisions by $64\%$. However, this approach also results in reduced average robot speed and increased unsmoothness, highlighting the need for a more sophisticated mechanism that can provide additional safety while also maintaining effectiveness.
The planning-based collision avoidance mechanism $\mathsf{DWA}$ effectively reduces the number of actual collisions and maximum braking scenarios, while also improving average robot speed by $42\%$ and reducing unsmoothness by $25\%$. However, it requires additional computational overhead due to the planning process. We then extend our analysis to compare traditional trajectory search, $\mathsf{DWA}$, with a learned policy, $\mathsf{RL}$. While the latter approach significantly reduces computational overhead by up to $85\%$ with improved average robot speed, it also compromises the safety of the system with a $50\%$ increase in actual collisions. These findings motivate our proposed approach SAFER, which demonstrates fewer actual collisions and dangerous scenarios than $\mathsf{DWA}$ and $\mathsf{RL}$, improves average robot speed compared to $\mathsf{DWA}$, and reduces unsmoothness compared to $\mathsf{RL}$. Moreover, the increase in computational overhead compared to $\mathsf{RL}$ is small, and SAFER is faster than $\mathsf{DWA}$.

\underline{Intentional Collision}
To further validate our approach compared to the baselines, we performed a challenging and unique experiment where we tasked a human operator with intentionally crashing the robot randomly in the evaluation environment for ten minutes. We show the results in Table~\ref{tbl:humanexp}. From the table, our approach $\mathsf{SAFER}$ is able to achieve zero actual crashes, and is able to avoid maximum braking conditions far better than all of the baselines. Additionally, the average speed is higher, and the unsmoothness is lower. We also include the average cost of actions taken by each baseline during their collision avoidance cycles, according to Eq.~\ref{eq:cost}. Our approach selects the most optimal control actions to avoid collision in comparison to all of the baselines. We believe that due to the fast, and more optimal control of our collision avoidance method, it can perform very well even under unpredictable driving behavior.

\begin{figure*}[t!]
    \centering
    \vspace{-2mm}
    \includegraphics[width=0.90\linewidth]{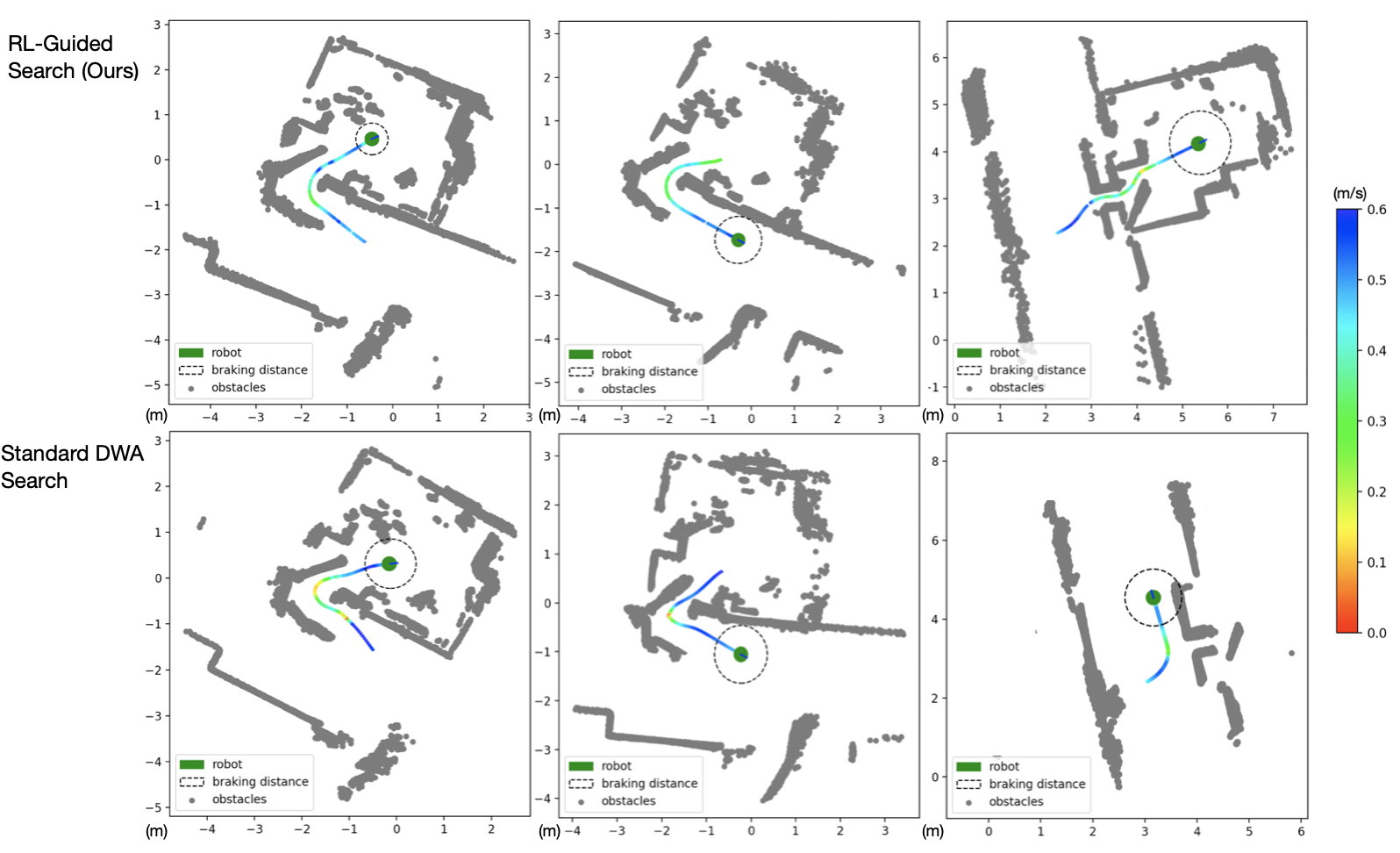}
    \vspace{-2mm}
    \caption{Qualitative comparison of our RL-Guided Search in top row, and Standard DWA Search in bottom row. A simple driving policy that applies maximum forward throttle (constantly) was used to try to drive the robot through tight doorways surrounded by obstacles, and the resulting path, and speeds are plotted for each obstacle avoidance method. In the first two scenarios, our method produces smoother collision avoidance paths without sudden reduction in speed (maximum braking) as in the standard search. In the last scenario, our method is able to effectively plan a path through the challenging doorway while the standard search is not and veers away.
    }
    \vspace{-6mm}
    \label{fig:qualitative}
\end{figure*}

\subsection{Qualitative Discussion}
In Fig.~\ref{fig:qualitative} we show qualitative results comparing the collision avoidance paths between our RL-guided search (top row), and standard DWA search (bottom row). We plot the speed of the robot along the path according to a color-map, and show the obstacle scan from the lidar sensor. In each of these scenarios, we mark a starting location and angle for the robot, and use a simple driving policy that applies maximum forward throttle and no turn. We repeat these scenarios ten times for each method, and select the best-performing trial for each. In the first two columns, the robot must traverse a sharp corner to turn through a doorway, and enter a cluttered room. The standard DWA search performed hard braking twice in the first column, and once in the second, whereas our RL-guided focused search performed zero braking. We also observe that the collision avoidance path in our method is smoother than that in the standard search. In the last column, the robot is presented with a narrow doorway where multiple turns are required to squeeze into the room. As can be seen from the images, our RL-guided focused search successfully makes it into the room, requiring only one slowdown in speed. The standard search is unable to plan an effective path, and turns away from the narrow passage, never discovering the room. We attribute this to the smaller search granularity that our method has compared to standard search.

\vspace{-1mm}
\section{Conclusion}
\vspace{-1mm}
In this work, we leverage the activation of emergency interventions as a learning signal for a real-world RL-guided search method that efficiently and effectively avoids collision. Our approach enables direct and safe real-world learning without human intervention. In our real-world experiments, we show our method outperforms other collision avoidance baselines. For future work, we plan to extend our setting to non-stationary multi-agent environments with potentially diverse dynamic obstacles. 

{
\small
\bibliography{ref_icra_2023}
\bibliographystyle{plain}
}

\end{document}